%% file: main.tex
\documentclass[10pt,letterpaper,twocolumn]{article}
\usepackage[latin9]{inputenc}
\usepackage{array}
\usepackage{soul}
\usepackage{multirow}
\usepackage{amsmath}
\usepackage{amssymb}
\usepackage{graphicx}
\usepackage{MnSymbol}
\usepackage[dvipsnames]{xcolor}
\usepackage[unicode=true,
 bookmarks=false,
 breaklinks=true,pdfborder={0 0 1},backref=section,colorlinks=false]
 {hyperref}
\hypersetup{
 pagebackref=true,letterpaper=true,colorlinks}

\makeatletter

\providecommand{\tabularnewline}{\\}
\newcommand{\lyxdot}{.}

\usepackage{cvpr}
\usepackage{times}
\usepackage{epsfig}
\usepackage{graphicx}

\cvprfinalcopy 

\def\cvprPaperID{****} 

\ifcvprfinal\fi

\makeatother

\begin{document}

\title{UR2KiD: Unifying Retrieval, Keypoint Detection, and Keypoint Description
without Local Correspondence Supervision}
\author{{Tsun-Yi Yang\thanks{Equal contribution. This work was mostly done in Scape Technologies research internship period.} $^{,1,2}$ \hspace{20pt} Duy-Kien Nguyen\footnotemark[1] $^{,1,3}$
\hspace{20pt} Huub Heijnen$^{1}$ \hspace{20pt} Vassileios Balntas$^{1}$}\\
{$^{1}$Scape Technologies \hspace{40pt} $^{2}$National
Taiwan University \hspace{40pt} $^{3}$Tohoku University}\\
\texttt{{\scriptsize{}tsun-yi@scape.io}}{\scriptsize{}\hspace{10pt}}\texttt{{\scriptsize{}
kien@vision.is.tohoku.ac.jp}}{\scriptsize{} \hspace{10pt}}\texttt{{\scriptsize{}huub@scape.io}}{\scriptsize{}
\hspace{10pt} }\texttt{{\scriptsize{}vassileios@scape.io}}{\scriptsize{}\hspace{5pt}}\texttt{{\scriptsize{} }}}

\maketitle
\input{abstract.tex}\input{introduction.tex}

\input{related_work.tex}

\input{method.tex}\input{experiments.tex}\input{conclusion.tex}

{\small{}\bibliographystyle{ieee_fullname}
\bibliography{UR2KID_ref}
 }{\small\par}
 
\end{document}

%% file: abstract.tex
\begin{abstract}
In this paper, we explore how three related tasks, namely keypoint detection, description, and image retrieval can be jointly tackled using a single unified framework, which is trained without the need of training data with point to point correspondences. By leveraging diverse information from sequential layers of a standard ResNet-based architecture, we are able to extract keypoints and descriptors that encode local information using generic techniques such as local activation norms, channel grouping and dropping, and self-distillation. Subsequently, global information for image retrieval is encoded in an end-to-end pipeline, based on pooling of the aforementioned local responses. In contrast to previous methods in local matching, our method does not depend on pointwise/pixelwise correspondences, and requires no such supervision at all i.e. no depth-maps from an SfM model nor manually created synthetic affine transformations. We illustrate that this simple and direct paradigm, is able to achieve very competitive results against the state-of-the-art methods in various challenging benchmark conditions such as viewpoint changes, scale changes, and day-night shifting localization. 
\end{abstract}


%% file: introduction.tex
\section{Introduction}

\begin{figure}
\begin{centering}
\begin{tabular}{cc}
\includegraphics[scale=0.3]{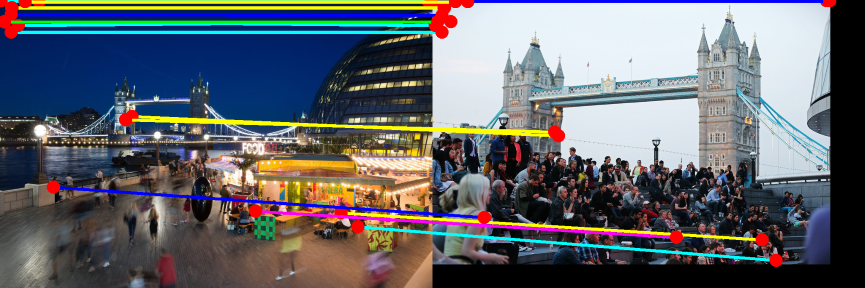} & \tabularnewline
(a) Pre-trained ResNet101  & \tabularnewline
\includegraphics[scale=0.3]{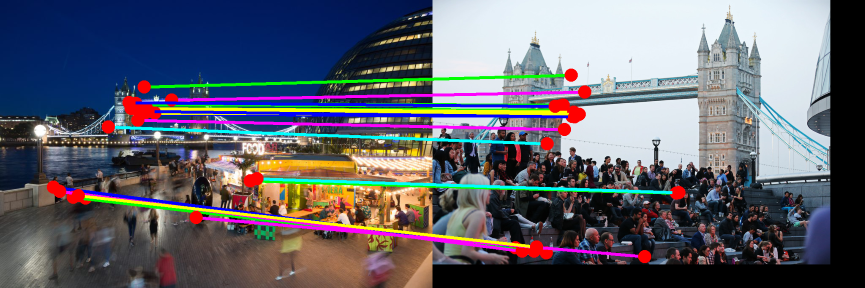} & \tabularnewline
(b) D2-Net \cite{dusmanu2019d2} & \tabularnewline
\includegraphics[scale=0.3]{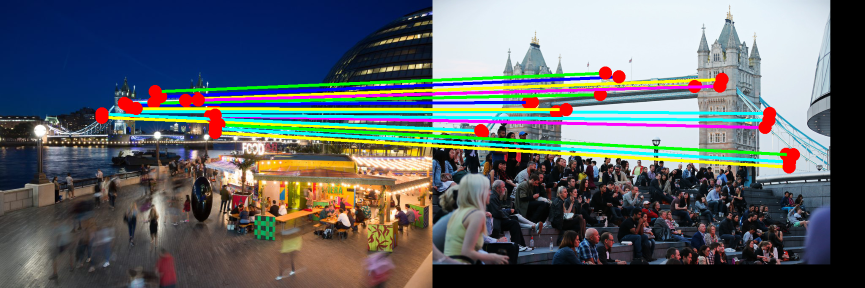} & \tabularnewline
(c) {\bf UR2KID (ours)} & \tabularnewline
\end{tabular}
\par\end{centering}
\caption{\label{fig:example}Extremely challenging image matching scenario with severe scale change and significant scene difference between day and night. The proposed UR2KID method is able to utilize a common network structure to achieve state-of-the-art results.}
\end{figure}

Image matching is one of the most important research
topics in computer vision with several applications such as 3D reconstruction \cite{schoenberger2016sfm,schoenberger2016mvs,theia-manual,sweeney2016large}, visual tracking \cite{gauglitz2011evaluation,yang2011recent,kim2015sowp} and SLAM \cite{mur2015orb,mur2017orb}. Several hand-crafted feature descriptors have been proposed \cite{Lowe2004Distinctive,Yang2016Accumulated,rublee2011orb,alahi2012freak,leutenegger2011brisk,balntas2015bold}, and have been widely utilized in state-of-the-art systems.
Similarly to other areas of computer vision, deep learning has recently influenced this area, with a plethora of works focusing on learning deep patch descriptors in a supervised manner \cite{tian2017L2,tian2019sosnet,balntas2016pn,yang2017deepcd,balntas2016learning,mishchuk2017working,he2018local,mukundan2019explicit} exhibiting superior performance compared to the ``classical'' hand-crafted methods.

Recently, an important direction of research with significant impact has been combining the whole matching pipeline into a single end-to-end process 
\cite{dusmanu2019d2,ono2018lf,yi2016lift,detone2018superpoint}. This enables the network to take advantage of several nuance factors that are related and contribute to matching. The detect-and-describe concept \cite{dusmanu2019d2} jointly encodes the keypoints and the descriptors in a single feature map. However, a significant drawback of such methods is that the training process is strongly supervised either by pixel-level correspondences from dense Structure-from-Motion (SfM) models which are extremely costly to produce~\cite{li2018megadepth}, or the application of manual synthetic affine transformations which unfortunately do not exhibit all the complex deformations seen in the real world \cite{detone2018superpoint}.

In very large-scale applications such as city-scale localization, exhaustive
brute force matching of all possible image pairs with local descriptors is an extremely 
costly and non-scalable process. An intuitive way to limit the candidate pool of images, thus making the problem tractable, is to utilise global image retrieval to limit the search space, and subsequently perform re-ranking using more accurate image matching and geometric verification methods \cite{sarlin2019coarse,sattler2018benchmarking}. 

 Nevertheless, previous studies focused on separate concepts for training or optimizing local matching pipeline and global descriptor retrieval as several independent parts. For instance, local matching methods are normally evaluated using local correspondences, and are frequently split into evaluation of keypoint robustness \cite{laguna2019key} and descriptor matching performance \cite{balntas2019hpatches}. However, such local methods neglect global context information that is by definition encoded into global descriptors for the task of image retrieval. Our key observation is that the tasks of image retrieval and image matching are highly interconnected, and a suitable optimization process can tackle both. For example, parts of the image that are suitable for global description such as buildings and static structures, would also normally be suitable for local matching. On the other hand, people, trees and cars, are normally unsuitable for both problems. Despite their seemingly obvious relation, these tasks have been
tackled either completely independently \cite{tian2017L2,tian2019sosnet,laguna2019key, Radenovi2016CNN}, or with minimal interaction
\cite{Noh2017Large,sarlin2019coarse}.

\begin{table*}
\begin{centering}
{\small{}{}}%
\begin{tabular}{|c|c||c|c|c|c|c|c|}
\hline 
\multirow{1}{*}{{\footnotesize{}{}Method}} & \multicolumn{1}{c|}{\multirow{1}{*}{{\footnotesize{}{}Training dataset}}} & \multicolumn{3}{c|}{{\footnotesize{}{}Ground truth label}} & \multicolumn{2}{c|}{{\footnotesize{}{}Local}} & {\footnotesize{}{}Global}\tabularnewline
\hline 
 &  & {\footnotesize{}{}How to get}  & {\footnotesize{}{}Img pair}  & {\footnotesize{}{}Loc. corres}  & {\footnotesize{}{}Detector}  & {\footnotesize{}{}Descriptor}  & {\footnotesize{}{}Descriptor}\tabularnewline
\hline 
\hline 
\multicolumn{8}{|c|}{\textbf{\footnotesize{}{}Keypoint detection}}\tabularnewline
\hline 
{\footnotesize{}{}QuadNet}  & {\footnotesize{}{}DTU robot image dataset}  & {\footnotesize{}{}3D}  & {\footnotesize{}{}-}  & {\footnotesize{}{}$\checkmark$}  & {\footnotesize{}{}$\checkmark$}  & {\footnotesize{}{}-}  & {\footnotesize{}{}-}\tabularnewline
\hline 
{\footnotesize{}{}Key.Net}  & {\footnotesize{}{}ImageNet ILSVRC 2012}  & {\footnotesize{}{}Self}  & {\footnotesize{}{}$\checkmark$}  & {\footnotesize{}{}-}  & {\footnotesize{}{}$\checkmark$}  & {\footnotesize{}{}-}  & {\footnotesize{}{}-}\tabularnewline
\hline 
\multicolumn{8}{|c|}{\textbf{\footnotesize{}{}Descriptor learning}}\tabularnewline
\hline 
{\footnotesize{}{}HardNet}  & {\footnotesize{}{}UBC/Brown dataset}  & {\footnotesize{}{}MVS}  & {\footnotesize{}{}-}  & {\footnotesize{}{}$\checkmark$}  & {\footnotesize{}{}-}  & {\footnotesize{}{}$\checkmark$}  & {\footnotesize{}{}-}\tabularnewline
\hline 
{\footnotesize{}{}SOSNet}  & {\footnotesize{}{}UBC/Brown dataset}  & {\footnotesize{}{}MVS}  & {\footnotesize{}{}-}  & {\footnotesize{}{}$\checkmark$}  & {\footnotesize{}{}-}  & {\footnotesize{}{}$\checkmark$}  & {\footnotesize{}{}-}\tabularnewline
\hline 
\hline 
\multicolumn{8}{|c|}{\textbf{\footnotesize{}{}Matching pipeline}}\tabularnewline
\hline 
{\footnotesize{}{}LIFT}  & {\footnotesize{}{}Piccadilly Circus dataset}  & {\footnotesize{}{}SfM}  & {\footnotesize{}{}$\checkmark$}  & {\footnotesize{}{}$\checkmark$}  & {\footnotesize{}{}$\checkmark$}  & {\footnotesize{}{}$\checkmark$}  & {\footnotesize{}{}-}\tabularnewline
\hline 
{\footnotesize{}{}Superpoint}  & {\footnotesize{}{}MS-COCO}  & {\footnotesize{}{}Self}  & {\footnotesize{}{}$\checkmark$} & {\footnotesize{}{}$\checkmark$}  & {\footnotesize{}{}$\checkmark$}  & {\footnotesize{}{}$\checkmark$}  & {\footnotesize{}{}-}\tabularnewline
\hline 
\multirow{1}{*}{{\footnotesize{}{}LF-Net}} & {\footnotesize{}{}ScanNet, 25 photo-tourism}  & {\footnotesize{}{}SfM}  & \multirow{1}{*}{{\footnotesize{}{}$\checkmark$}} & \multirow{1}{*}{{\footnotesize{}{}$\checkmark$} } & {\footnotesize{}{}$\checkmark$}  & {\footnotesize{}{}$\checkmark$}  & {\footnotesize{}{}-}\tabularnewline
\hline 
{\footnotesize{}{}D2-Net}  & {\footnotesize{}{}MegaDepth dataset}  & {\footnotesize{}{}SfM}  & {\footnotesize{}{}$\checkmark$}  & {\footnotesize{}{}$\checkmark$}  & {\footnotesize{}{}$\checkmark$}  & {\footnotesize{}{}$\checkmark$}  & {\footnotesize{}{}-}\tabularnewline
\hline 
{\footnotesize{}{}R2D2}  & {\footnotesize{}{}Oxford, Paris, Aachen (scene specific)}  & {\footnotesize{}{}SfM, Flow, Style}  & {\footnotesize{}{}$\checkmark$}  & {\footnotesize{}{}$\checkmark$}  & {\footnotesize{}{}$\checkmark$}  & {\footnotesize{}{}$\checkmark$}  & {\footnotesize{}{}-}\tabularnewline
\hline 
{\footnotesize{}{}ELF}  & {\footnotesize{}{}ImageNet pre-trained}  & {\footnotesize{}{}-}  & {\footnotesize{}{}-}  & {\footnotesize{}{}-}  & {\scriptsize{}{}$\filledmedtriangleup$}  & {\scriptsize{}{}$\filledmedtriangleup$}  & {\footnotesize{}{}-}\tabularnewline
\hline 
\hline 
\multicolumn{8}{|c|}{\textbf{\footnotesize{}{}Retrieval}}\tabularnewline
\hline 
{\footnotesize{}{}NetVLAD}  & {\footnotesize{}{}Google street view}  & {\footnotesize{}{}T.M.}  & {\footnotesize{}{}$\checkmark$}  & {\footnotesize{}{}-}  & {\footnotesize{}{}-}  & {\footnotesize{}{}-}  & {\footnotesize{}{}$\checkmark$}\tabularnewline
\hline 
{\footnotesize{}{}GeM, DAME}  & {\footnotesize{}{}SfM-120k (cluster+3D)}  & {\footnotesize{}{}SfM}  & {\footnotesize{}{}$\checkmark$}  & {\footnotesize{}{}-}  & {\footnotesize{}{}-}  & {\footnotesize{}{}-}  & {\footnotesize{}{}$\checkmark$}\tabularnewline
\hline 
\hline 
\multicolumn{8}{|c|}{\textbf{\footnotesize{}{}Multi-task method}}\tabularnewline
\hline 
{\footnotesize{}{}DELF}  & {\footnotesize{}{}Landmark dataset}  & {\footnotesize{}{}Class} & {\footnotesize{}{}-} & {\footnotesize{}{}-} & {\scriptsize{}{}$\filledmedtriangleup$} & {\scriptsize{}{}$\filledmedtriangleup$}  & {\footnotesize{}{}$\checkmark$}\tabularnewline
\hline 
{\footnotesize{}{}HF-Net} & {\footnotesize{}{}Google landmark, BDD}  & {\footnotesize{}{}Teacher}  & {\footnotesize{}{}$\checkmark$}  & {\footnotesize{}{}$\checkmark$}  & {\footnotesize{}{}$\checkmark$} & {\footnotesize{}{}$\checkmark$}  & {\footnotesize{}{}$\checkmark$}\tabularnewline
\hline 
{\footnotesize{}{}ContextDesc} & {\footnotesize{}{}Photo-tourism, aerial dataset} & {\footnotesize{}{}SfM} & {\footnotesize{}{}$\checkmark$} & {\footnotesize{}{}$\checkmark$} & {\footnotesize{}{}$\checkmark$} & {\footnotesize{}{}$\checkmark$}  & {\scriptsize{}{}$\filledmedtriangleup$}\tabularnewline
\hline 
{\footnotesize{}{}UR2KID}  & {\footnotesize{}{}ImageNet pre-trained, MegaDepth}  & {\footnotesize{}{}SfM}  & {\footnotesize{}{}$\checkmark$}  & {\footnotesize{}{}-}  & {\footnotesize{}{}$\checkmark$}  & {\footnotesize{}{}$\checkmark$}  & {\footnotesize{}{}$\checkmark$}\tabularnewline
\hline 
\end{tabular}
\par\end{centering}
\begin{centering}
 
\par\end{centering}
\caption{\label{tab:Overall-comparison}Overall comparison of related methods.  {\scriptsize{}{}$\filledmedtriangleup$} indicates that this specific task is not directly optimized by this method.}
\end{table*}


In this paper, we propose to unify feature encoding in terms of both local and
global information, using a multi-task learning approach. Unlike traditional
patch based methods, our local matching feature maps are suitable for both
keypoint detection and description. Multi-task information is embedded in a single network which is also responsible for learning a global descriptor suitable for image retrieval. We focus on the following contributions:


\begin{itemize}
  \setlength\itemsep{0.1em}
    \item We present a multi-task method for global
    retrieval and local matching embedded within a single network with a training process that does not rely on local pixel level correspondence ground truth.
    \item We introduce a novel method to aggregate feature map descriptors, namely group-concept detect-and-describe (GC-DAD).
    Using our method, we show that a standard deep network trained on ImageNet for a classification task can match or outperform state-of-the-art end-to-end trained matching methods.
    \item We show that by using cross dimensional self-distillation along with the proposed training method with no pixel-level correspondences, our method is able to acquire low dimensional local descriptor which is more robust against scale changes, viewpoint changes, and day-night shifting localization as shown in Figure.\ref{fig:example}. Our combined global and local descriptor, is able to outperforms state-of-the-art localization methods.
\end{itemize}{}

%% file: related_work.tex
\section{Related Work}

In this section, we briefly discuss the classical patch based matching pipeline, the state-of-the-art end-to-end image matching methods, and the relation between local and global matching. 

\textbf{Classical matching pipeline.} The classical matching pipeline can be deconstructed in the following components: keypoint detection, description, and matching. 
\textbf{a) Keypoint detection}: For sparse matching, finding robust keypoints is the first crucial part.
Several properties should be concurrently satisfied, such as scale \& affine invariance, and keypoint repeatability \cite{mishkin2018repeatability,mikolajczyk2002affine,Lowe2004Distinctive,moo2016learning,rublee2011orb,moo2016learning,laguna2019key,revaud2019r2d2}.
\textbf{b) Keypoint description}: Given a keypoint location and a corresponding scale associated with it, a patch can be rectified around it, and subsequently described either by hand-crafted method such as SIFT \cite{Lowe2004Distinctive}, SURF \cite{bay2006surf,Bay2008Speed}, LIOP \cite{wang2011local}, ORB \cite{rublee2011orb,mur2015orb,mur2017orb} or by deep learned patch description methods such as \cite{tian2017L2,tian2019sosnet,balntas2016pn,yang2017deepcd,balntas2016learning,mishchuk2017working,he2018local,mukundan2019explicit}.



\textbf{End-to-end matching pipelines.} 
In the classical pipeline, the main individual components (i.e. keypoint detection and description) are often tackled separately which might lead to sub-optimal results. 
Recent approaches try to tackle this issue by introducing end-to-end matching pipelines.
For preserving the differentiability over the whole matching pipeline, the authors of LIFT \cite{yi2016lift} replace the keypoint detection and affine estimation by a spatial transformer network \cite{jaderberg2015spatial} and the non-maxima suppression by a soft argmax. 
By using a dense SfM model during training, viewpoint and lighting conditions are resolved by training a siamese network. 
Similarly, Lf-Net \cite{ono2018lf} and Superpoint \cite{detone2018superpoint} use synthetic affine transformations for generating image pairs as training data, which allows the network learning to be based on point-wise pixel level ground truth local correspondences. 
D2-Net \cite{dusmanu2019d2} is also trained based on pointwise ground-truth from a set of dense SfM models \cite{li2018megadepth}, and its main contribution, was the proposed detect-and-describe process which entails using same feature map for keypoint detection and description. 
ELF \cite{benbihi2019elf} exploits the possibility of pre-trained network through back-propagated saliency map to get robust detector and matching results.

\textbf{Retrieval methods.} 
Where feature matching uses pixel level descriptions and correspondences, image retrieval uses a single (global) description for the whole image.
\cite{Teichmann2019Detect,Babenko2014Neural,Radenovi2018Revisiting,Radenovi2018Fine,Radenovi2016CNN,Arandjelovic2016NetVLAD,Xu2018Unsupervised,Xu2015Unsupervised}. 
The similarity between two images can then be computed very efficiently with a single distance computation, something that makes the process useful for retrieving a pool of possible matching candidates in large-scale dataset, with respect to a given query image. 
Recent state-of-the-art methods aggregate feature maps into global descriptors using an $\ell_{3}$-norm in Generalized-Mean (GeM) \cite{Radenovi2018Fine} and more recently a learnable dynamic $\ell_{p}$-norm in DAME \cite{yang2019dame}.
Detect-to-retrieve \cite{Teichmann2019Detect} adopts an extra detection module for sub-region feature extraction before the final aggregation.

\textbf{Multi-task methods.} 
Considering both local correspondence and global description aim to induce a similarity metric between two images, joint learning local matching and global retrieval might be beneficial for both sides. 
For the purpose of landmark classification using global descriptors, local masks and image pyramids are adopted in DELF \cite{Noh2017Large}. 
Here, even though local matching is explored as a side-product, it is not directly optimized for the task.
HF-Net \cite{sarlin2019coarse} adopts a teacher-student distillation for both matching and retrieval in order to achieve fast and robust localization. 
ContextDesc \cite{luo2019contextdesc} encodes both visual and geometric context along with the information from off-the-shelf retrieval network trained on Google-landmarks~\cite{Noh2017Large} into the description. 

In Table \ref{tab:Overall-comparison} we present a systematic categorization of related techniques with respect to training methodologies and method outcomes.
Despite the fact that some of these methods can be used for both matching and retrieval (e.g., DELF, HF-Net, Contextdesc) there is no simple unified framework for learning the both tasks concurrently. 
In the following section we will introduce the proposed UR2KiD method that aims to explore several problems related to learning a simple unified multi-task method that is suitable for both image matching and retrieval.

%% file: method.tex
\section{Methodology}

\begin{figure*}[t]
\begin{centering}
\begin{tabular}{cc}
\includegraphics[scale=0.5]{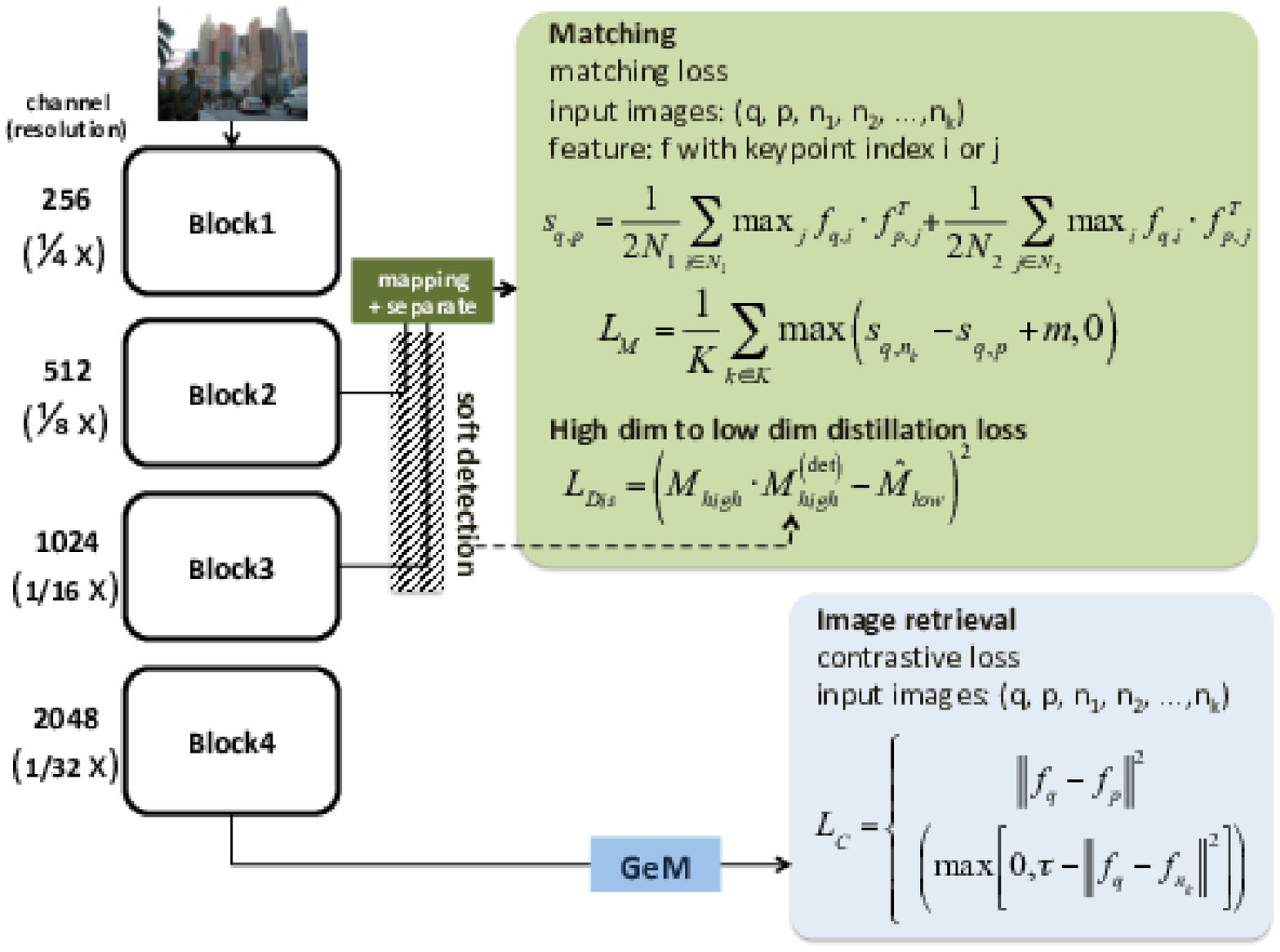}  & \includegraphics[scale=0.5]{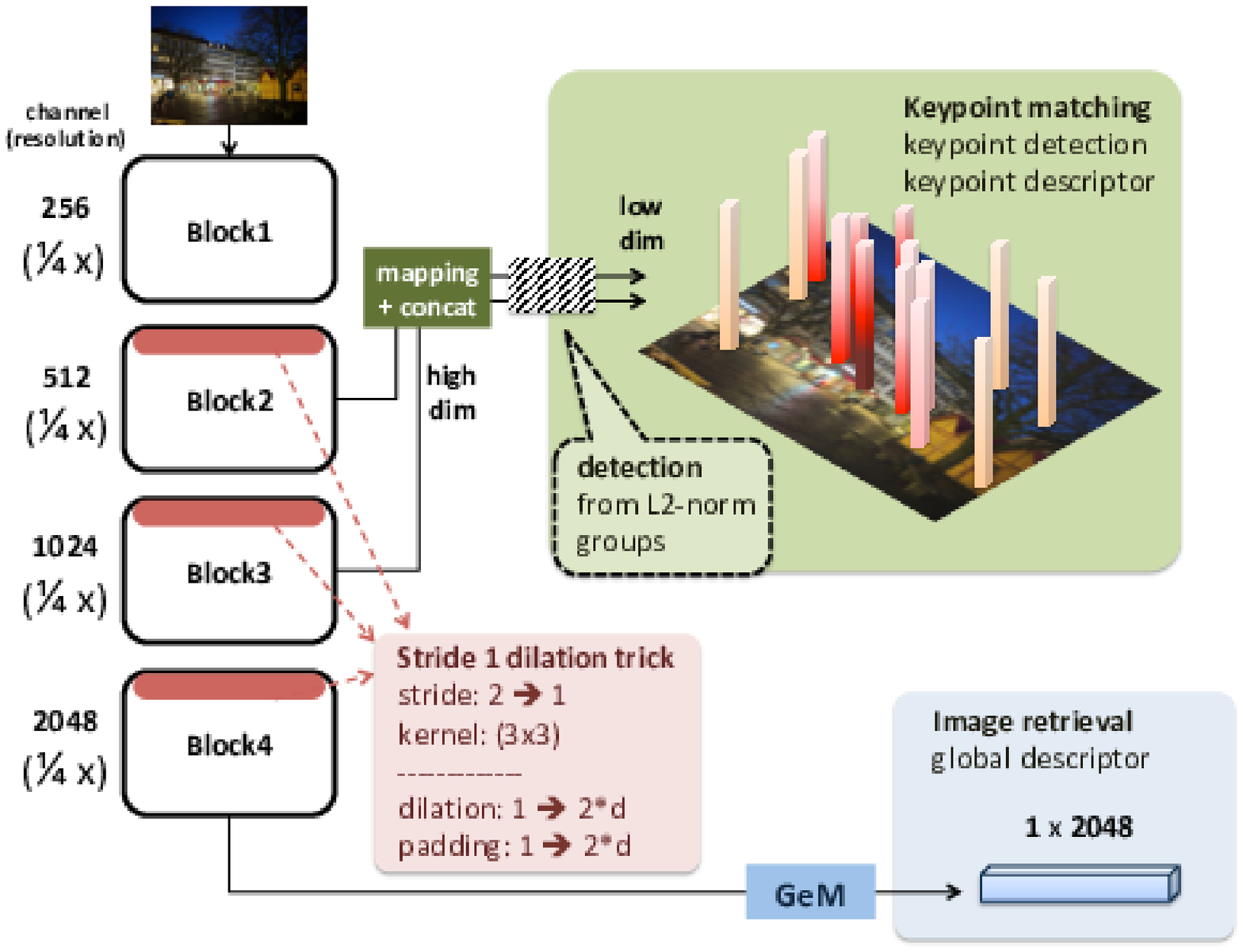}\tabularnewline
(a) Training  & (b) Testing\tabularnewline
\end{tabular}
\par\end{centering}
\caption{\label{fig:Overview}Overview of the proposed method.}
\end{figure*}

Given an image $I$, the extraction of the visual information into a suitable mathematical representations also referred to as visual descriptors, can be formulated in two
ways, either globally or locally. The global descriptors are extracted using the entire image, and are normally utilised in retrieval problems. The local descriptors, use information from small regions of the image, and are normally used
for image matching scenarios. 
For each image $I$, the output of these processes, is a set of keypoints $k_i$, a set of descriptors $d_i$ corresponding to index $i$, and a single global descriptor $d_{g}$. 



Unlike previous methods that treat local and global descriptors separately, our goal is to generate both global and local description within a single network pipeline $P_{joint}$ 
\begin{equation}
d_{g},\left\{ k_{i},d_{i}\right\} =P_{\text{joint}}\left(I\right).
\end{equation}
The given training supervision is the image level pair $\left(a,p,n_{1},n_{2},...,n_{k}\right)$
where $\left(a,p\right)$ is the positive pair and $\left(a,n_{k}\right)$
is the negative pairs without the pixelwise or pointwise matching
correspondences. The image pairs come from SfM model and the whole process does not depend on any human labeled ground truth.


We now briefly describe the overview of our method. In Figure.~\ref{fig:Overview}, the
training and testing pipelines are demonstrated, and we present a general visual outline of our method using blocks to represent different parts of the system. As a simplification, we adopt the block design of the ResNet for the high-level network explanation (e.g. capital B as Block. B1 for Block1.). Our goal is to utilize the feature map hierarchy of a convolutional neural network, which has known to contain different levels of semantics, and produce multi-level representations
of an image with discriminative features for image retrieval problem. We use a pretrained CNN (ResNet101 \cite{He2016Deep} trained on ImageNet \cite{deng2009imagenet,Russakovsky2015ImageNet}) to extract multi-level visual features.

The architecture is composed of the following components:  multi-level feature extraction for local descriptors from ResNet blocks and feature pyramid network (FPN) \cite{lin2017feature}. The final global descriptor is the concatenation of several different GeM \cite{Radenovi2018Fine} pooling results from different layers.

\subsection{Local keypoint and description}

For the local matching paradigm, we first start by formalising the the detect-and-describe (DAD) method introduced in D2-Net~\cite{dusmanu2019d2}. 
To determine the $i$-th keypoint coordinate $(x_{i},y_{i})$ out of feature map $F$, the keypoint confidence of such location is computed by thresholding the maximum response of the feature map $F_{(x_{i},y_{i})}^{c^{*}}$ across channel dimension (i.e. $c^{*} = \arg \max_{c}F_{(x_{i},y_{i})}^{c}$) along with local non-maximum suppression and edge confidence thresholding of Harris corner detector, and the second-order spatial displacement first described in \cite{Lowe2004Distinctive}.

Our method is mainly inspired by D2-Net~\cite{dusmanu2019d2} and Simeoni et al \cite{simeoni2019local} who propose a way to explore the activations of CNNs as keypoint detectors, and use the parameters of classical detectors such as Harris or MSER \cite{matas2004robust} on the activation to match two images channel by channel. However, our key difference with Simeoni et al \cite{simeoni2019local}, is that their method doesn't exploit the discriminative information for the descriptor which limits their method to only retrieval re-ranking. Based on the idea of independent matching for each channel, we refine the process by aggregating the concept from different channels as \textbf{Group-Concept Detect-and-Describe (GC-DAD)} for both keypoint detection and description.

Unlike D2-Net, which takes takes the maximum value across feature map channel, our
\label{sec:gc-dad} method depends on the L2 response as the importance of the keypoint.
For extracting different concepts, we would like to divide the feature map into different groups as independent concepts. The collection of the groups is represented by $\left\{ F\right\}^{g}$ with
$g$ as the group index. For example, if a feature map $F$ contains $K$ channels and we want to uniformly divide it into $G$ groups, then $F^{g=1}$ is corresponding to channel $1\sim\left\lfloor \frac{K}{G}\right\rfloor$, and $F^{g}$ is corresponding to channel $((g-1)*\left\lfloor \frac{K}{G}\right\rfloor+1)\sim(\left\lfloor \frac{K}{G}\right\rfloor*g)$.
We may compute the $i$-th keypoint location $(x_{i},y_{i})$ out of the L2 response map of feature map $F^{g}$ and combining the keypoint sets from every group by setting a L2 response threshold on the L2-norm of each group $F_{(x_{i},y_{i})}^{g}$ along with Harris edge threshold and the 2-nd order displacement similarly to~\cite{dusmanu2019d2}.

\begin{figure}
\begin{centering}
\includegraphics[scale=0.35]{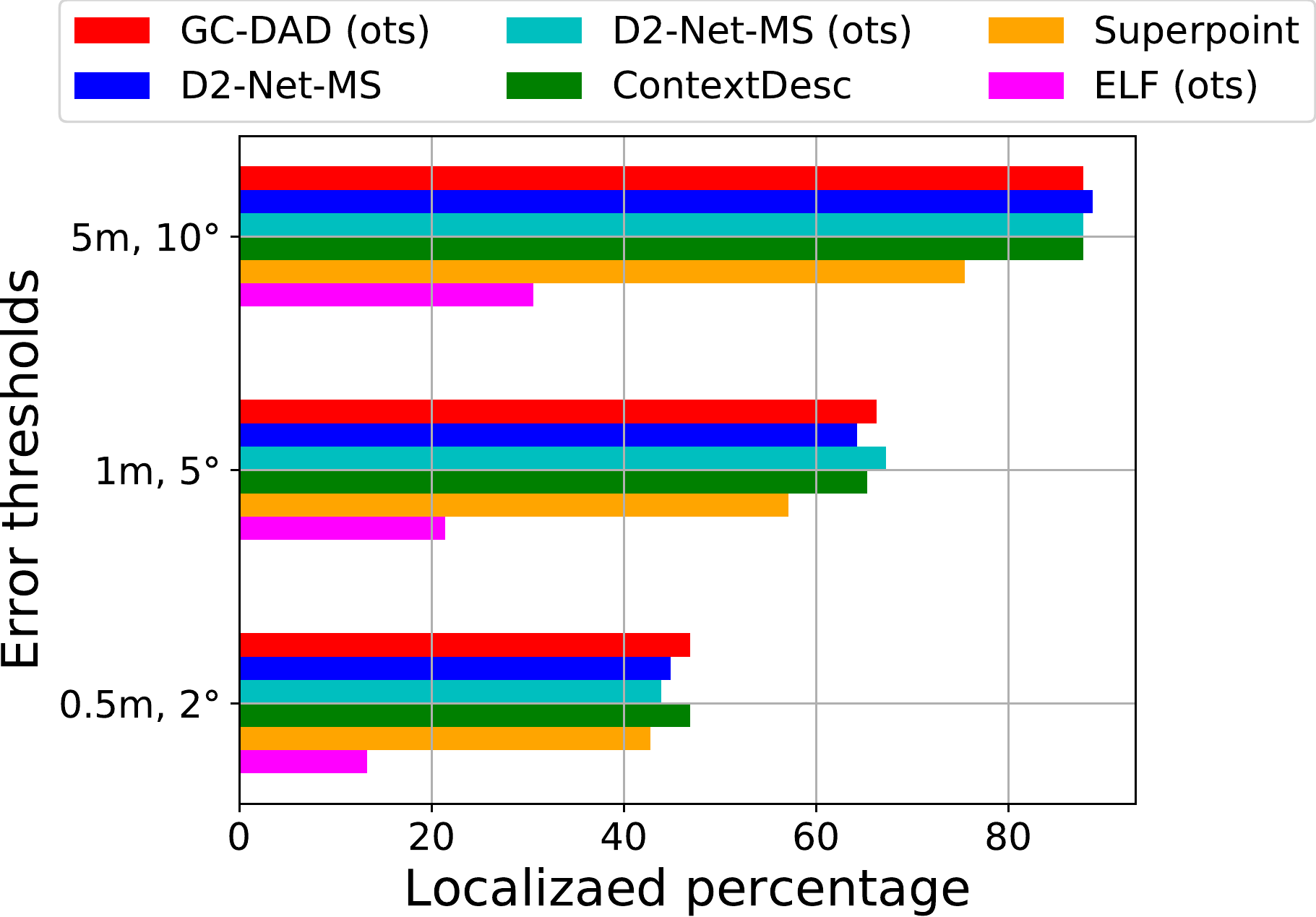} 
\par\end{centering}
\caption{\label{fig:bar_comparision}Aachen day-night localization benchmark \cite{sattler2018benchmarking} results. Comparison between 3 state of the art end-to-end matching methods, and our group-concept detect-and-describe paradigm (GC-DAD) with a simple pre-trained ImageNet ResNet101 network \cite{deng2009imagenet,Russakovsky2015ImageNet,He2016Deep}. Surprisingly, by carefully utilising the feature maps with the proposed GC-DAD method, even a pre-trained ImageNet ResNet101 is capable of defeating every state-of-the-art local matching method without any fine-tuning or further training.}
\end{figure}

To illustrate the power of the GC-DAD process, we show in Figure \ref{fig:bar_comparision} results for the Aachen dataset for the day-night shifting camera localization \cite{sattler2018benchmarking}. Our method is strong enough even by using a pretrained ImageNet off-the-shelf ResNet101 without any extra training, while others requires self-learning \cite{detone2018superpoint} or fine-tuning \cite{dusmanu2019d2}. 

There are some drawbacks of performing GC-DAD on high dimensional descriptors such as  descriptors aggregated from B2+B3 blocks of ResNet101 (more information on this will be discussed in the implementation details). The final descriptor is still memory costly since GC-DAD is only responsible for detecting the keypoints and the channel dimension out of the given feature map is preserved. In terms of practicality, a low dimensional descriptor is much more suitable
for large-scale applications. Apart from that, manually selected groups is not the optimal choice since there is no clear clue about how to aggregate them.
To that end, we explore a dimensionality reduction mapping with a channel dropping method in the next section.

\textbf{Concept dropout dimension reduction}
A straight forward way to implement the grouping would be dividing the channel into several parts evenly with the same dimension. However, each channel contains different information and concepts. There is no clear evidence about which channels
should be grouped together. Another trivial solution would be learning multiple mapping layers as aggregation learners for groups, but such setting is also redundant since several mapping layers have to defined for the groups.

To avoid the aforementioned problems of non-trivial grouping and high dimensional descriptor, we adopt 2D dropout on the feature maps
and perform a convolutional mapping for dimension reduction. By randomly dropping the feature channels in the training phase, it's conceptually selecting different
groups without manual selection, and brings the diverse concepts into
the matching and avoid overfitting.
\begin{equation}
\hat{F}=\text{Conv}\left(\text{Drop}_{\text{2D}}\left(F\right)\right)
\end{equation}
with feature map $F$ as the high dimensional input for the dimension reduction, and $\hat{F}$ as the low dimensional results.

\textbf{Matching loss for affinity matrix} For training the discriminative local
descriptor, given different input images $a,p$, and the corresponding
feature maps $\hat{F}_{a},\hat{F}_{p}$ along with the matching affinity matrix
$M$ can be computed by the inner product.
\begin{equation}
M = \hat{F}_{a}\cdot \hat{F}_{p}^{\mathrm{T}}
\label{eqn:affinity matrix}
\end{equation}
By taking the maximum value along the column and row of the matching
affinity matrix, the average score $s$ can be computed
\begin{equation}
s_{a,p} = \frac{1}{2N_{1}}\sum_{i\in N_{1}}\max_{j}\hat{f}_{a,i}\cdot \hat{f}_{p,j}^{\mathrm{T}} +\frac{1}{2N_{2}}\sum_{j\in N_{2}}\max_{i}\hat{f}_{a,i}\cdot \hat{f}_{p,j}^{\mathrm{T}}
\label{eqn:affinity score}
\end{equation}
with $i,j$ as the index of the column and row. A margin loss $L_{M}$ is adopted as
\begin{equation}
L_{M} = \frac{1}{K}\sum_{k\in K}\max\left(s_{a,n_{k}}-s_{a,p}+m,0\right).
\label{eqn:matching loss}
\end{equation}
which indicates the score of the positive pair is higher than the score of the negative pair by a margin $m$ in terms of matching. Note that optimzing the matching loss does not require pixel level nor pointwise correspondence supervision. The local supervision we
used is based on image pairs, similarly to the ones used for training image retrieval methods. This supervision is very weak comparing to the state-of-the-art
methods in Table \ref{tab:Overall-comparison}.

\begin{figure*}
\begin{centering}
\begin{tabular}{cccc}
\includegraphics[scale=0.3]{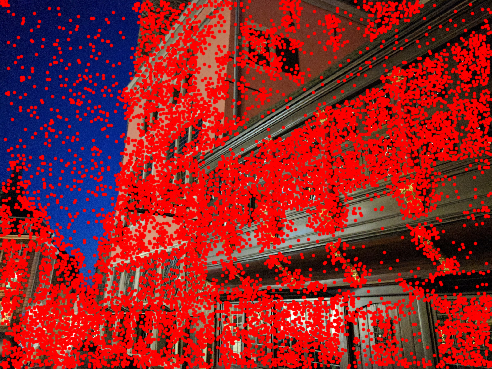}  & \includegraphics[scale=0.3]{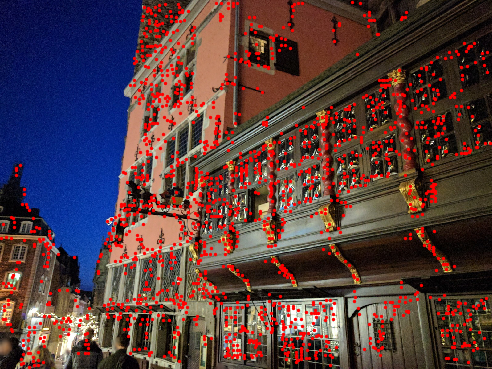}  & \includegraphics[scale=0.3]{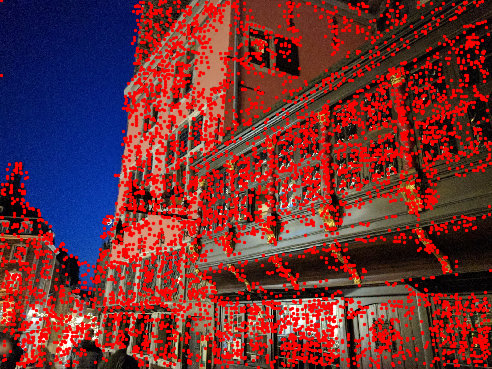}  & \includegraphics[scale=0.3]{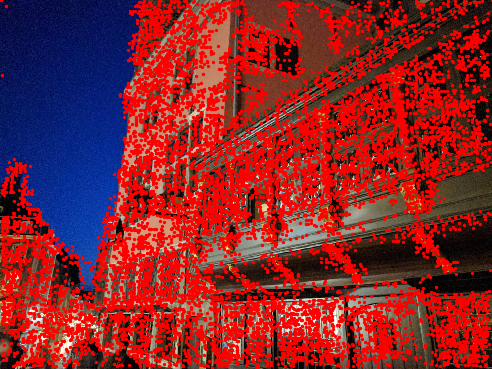}\tabularnewline
(a) D2-net  & (b) 1 group  & (c) 3 groups  & (d) 6 groups\tabularnewline
\end{tabular}
\par\end{centering}
\caption{\label{fig:Grouping-channel-keypoint}Grouping channel keypoint detection
comparison. D2-net chooses the maximum value of each channel which
results in high density of keypoints. On the other hand, our feature channel grouping technique
combined with the feature channel L2-norm  based thresholding can help us to control different levels of keypoint activation. Results here are based on the high dimensional teacher detection.}
\end{figure*}

\begin{figure*}
\begin{centering}
\begin{tabular}{cccc}
\includegraphics[scale=0.35]{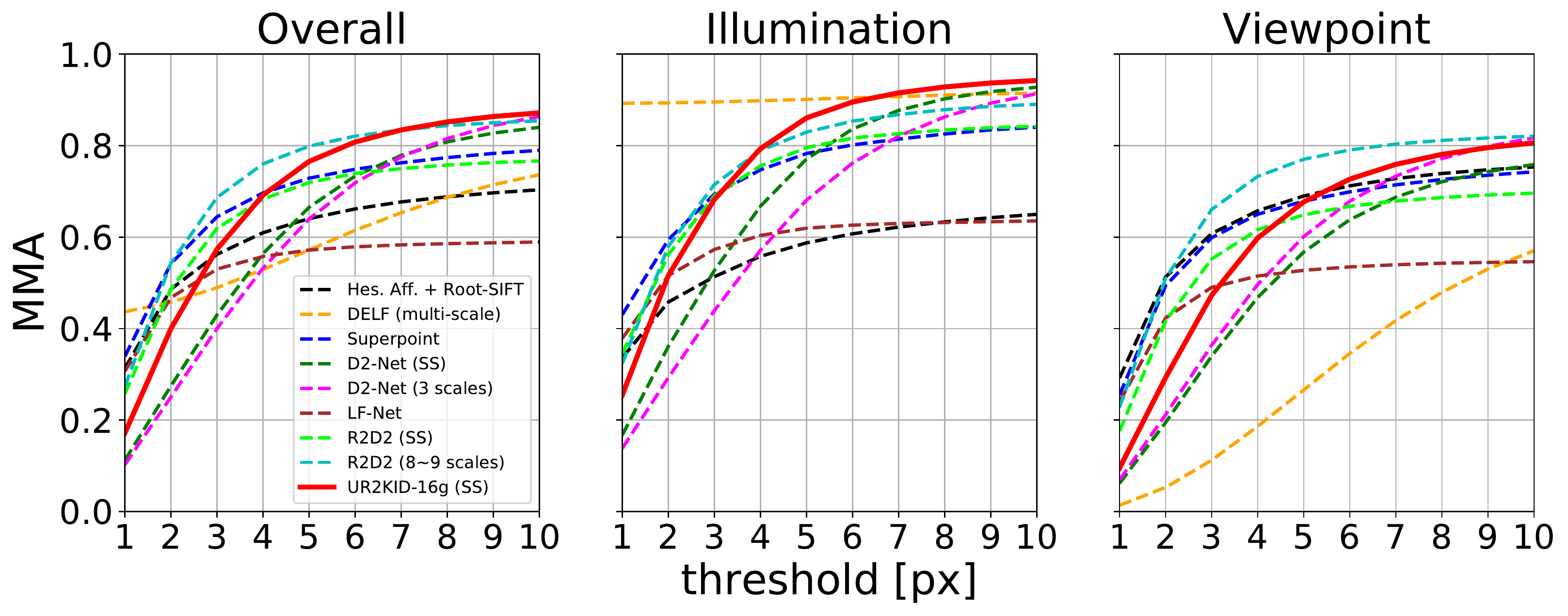} & \includegraphics[scale=0.35]{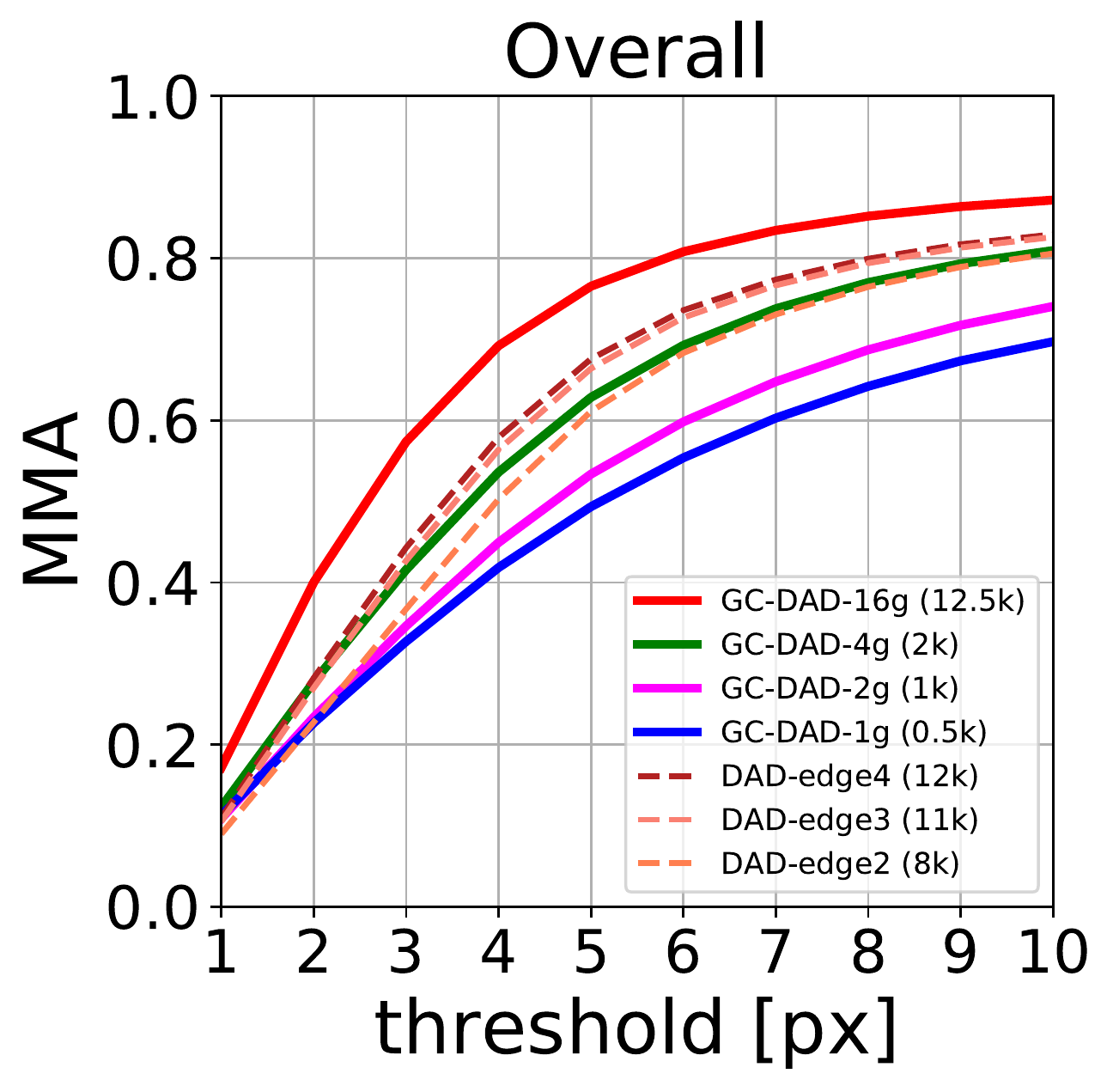} \tabularnewline
(a) Comparison  & (b) Ablation\tabularnewline
\end{tabular}
\par\end{centering}
\caption{\label{fig:hpatches} (a) Hpatches dataset mean matching accuracy (MMA) comparison. We can observe that the proposed method outperforms all other methods at the threshold of 4 pixels and above. (b) Ablation study for DAD and GC-DAD for the same network ResNet101. We can observe that more groups lead to increased performance while DAD has limited performance even with more keypoints (number in bracket).}
\end{figure*}

\textbf{Cross dimension distillation} During the feature mapping, the
low dimensional descriptor $\hat{F}$ cannot capture the whole information
from the high dimensional descriptor $F$ by only using a matching loss.
Normally, distillation \cite{hinton2015distilling, sarlin2019coarse} is a good way for transferring the information from a teacher model to a student model. Nevertheless,
HF-Net adopts direct distillation which requires same descriptor dimensions for
both teacher and student output, and in addition, the teacher models require training on extra dataset.

Here we propose a new way for distilling high dimension information
into a low dimensional descriptor indirectly. 
\begin{equation}
L_{Dis}=\left(M_{\text{high}}\cdot M_{\text{high}}^{\left(\text{det}\right)}-\hat{M}_{\text{low}}\right)^{2}
\end{equation}
We refer the high dimensional one as teacher and the low dimensional one as student.
The distillation happens on the matching affinity matrix $M_{\text{high}}$
and $\hat{M}_{low}$ with the corresponding soft detection score $M_{\text{high}}^{\left(\text{det}\right)}$
for the high dimensional feature $F$ as described in \cite{dusmanu2019d2}. The affinity matrix is irrelevant
to the feature dimension which is similar to \cite{tung2019similarity}. In order to avoid the training of low dimension descriptor affect the high dimension one, we adopt backbone freezing or gradient cutdown for the training for the distillation. Comparing
to HF-Net, our teacher descriptor is based on the same backbone with
concatenation trick (B2,B3), while HF-Net \cite{sarlin2019coarse} require extra teacher (DOAP \cite{he2018local},
NetVLAD \cite{Arandjelovic2016NetVLAD}) for the training.

\subsection{Global description}

\textbf{Generalized-mean pooling} Proposed in \cite{Radenovi2018Fine}, the global representation is given by the generalized mean operation (GeM) as
\begin{equation}
d_{g} =\left(\frac{1}{hw}\sum_{s\in hw}x_{c,s}^{p}\right)^{\frac{1}{p}}.
\label{eqn:GeM}
\end{equation}
with $hw$ as the total spatial dimension of the input feature map, $x_{c,s}$ as the feature from channel $c$ and spatial dimension $s$ where $s\in [0,hw]$ and $p=3$ as described in GeM \cite{Radenovi2018Fine}.
Following the previous works \cite{Radenovi2018Fine,dusmanu2019d2},
we use a siamese architecture and train a two-branch network. Each
branch shares the same network's architecture and parameters.

The final global descriptor is the concatenation of all the FPN output along with Eq.\ref{eqn:GeM} pooling.
To train the network on non-matching ($y=0$) and matching pairs ($y=1$), we employ
the contrastive loss.
\begin{equation}
L_{C}=\begin{cases}
\frac{1}{2}\|d_{g}(a)-d_{g}(p)\|^{2} & y=1\\
\frac{1}{2}(\max\{0,\tau-\|d_{g}(a)-d_{g}(n_{k})\|\})^{2} & y=0
\end{cases}
\label{eqn:Lc}
\end{equation}
where $\tau$ is the margin.

\subsection{Joint local and global training}

By considering the gloabal context and the local keypoint information, the final loss for jointly optimizing the local and global tasks is
\begin{equation}
L=L_{M}\left(F_{B2}\right)+L_{M}\left(F_{B3}\right)+L_{M}\left(\hat{F}\right)+L_{C}+\lambda L_{Dis}.
\label{eqn:L_all}
\end{equation}
Considering that the distillation loss $\lambda L_{Dis}$ is not directly optimizing the metric, we use a relatively small hyperparameter $\lambda=0.1$ parameter value for controlling it.

\subsection{Implementation details} 

For the feature map representation that is used, we feed a single-scale image of an arbitrary size, and output four feature maps at multiple levels, in a fully convolution style. We
do this to exploit all of the information from low to high level to
the global pooling method. We conjecture that different concept levels
will provide more useful features for a global image representation.

We first extract the features from the last four residual
blocks in bottom-up path way (after the ReLU and before the pooling
layers), which are in different sizes as described in Figure \ref{fig:Overview}.
Given the bottom-up feature map $C_{r}$ with layer $r$, the final representation we used for each layer is $F_{r}$ as described in the following equation.
\begin{equation}
F_{r} =\,\text{ReLU}\big(\text{Conv}_{3\times3}(C_{r}) +\text{Upsampling}(F_{r+1})\big)
\end{equation}
which follows the top-down merging in FPN \cite{lin2017feature}. The ReLU is applied to each feature map to make sure that it is non-negative.

Our local feature representation is made by
the B2 and B3 residual blocks as described in Figure \ref{fig:Overview} (B refers to a
Block) which are optimized jointly in training stage. During testing, in order to maintain multi-scale information, we take the feature maps from B2 \& B3 and concatenate them. The spatial differences between B2 and B3 causing by pooling are resolved by the
dilation trick in Figure \ref{fig:Overview} (b) which replaces stride
2 by stride 1 along with dilated convolution kernel. High resolution
of the output feature map is essential for keypoint detection and
matching.

%% file: experiments.tex
\section{Experiment}

In this section, we present several experimental results that illustrate the 
power of the proposed GC-DAD method, In addition, we present ablation studies related to several parameters of our method.

\subsection{Dataset and Setting}


\textbf{Training} Megadepth \cite{li2018megadepth} dataset was adopted for training which contains 1,070,468 images from 196 different scenes and reconstructed by COLMAP \cite{schoenberger2016sfm,schoenberger2016mvs} along with their depth maps and intrinsics / extrinsics matrices.

We use the Adam optimizer in training with the parameters $\alpha,\beta=(0.9,0.99)$. During the training procedure, we make the learning rate $\alpha$ decay at $i$-th epoch with an exponential rate of $\exp^{-0.1i}$. 
We treat each training sample as a tuple of one query, one positive and five negative images.


For the training, we use exactly the same image pairs as D2-Net \cite{dusmanu2019d2} from Megadepth dataset \cite{li2018megadepth} for the sake of fair comparison. However, we do not utilise the pointwise correspondences, or the depth map ground truth for our training method. 
All models are trained up to 100 epochs. The batch size is set to 5 tuples, and the margin $\tau$ is set to 0.85. For each training epoch, around 6k tuples are selected. 

\begin{table}
\begin{centering}
{\small{}}%
\begin{tabular}{c|c|c}
\hline 
 & \multicolumn{1}{c|}{{\footnotesize{}day}} & \multicolumn{1}{c}{{\footnotesize{}night}}\tabularnewline
\hline 
\multicolumn{3}{c}{{\footnotesize{}Protocal 1 (Pre-defined query candidates)}}\tabularnewline
\hline 
{\footnotesize{}ELF} & {\footnotesize{}-} & {\footnotesize{}13.3 / 21.4 / 30.6}\tabularnewline
\hline 
{\footnotesize{}SuperPoint} & {\footnotesize{}-} & {\footnotesize{}42.8 / 57.1 / 75.5}\tabularnewline
\hline 
{\footnotesize{}DELF (new)} & {\footnotesize{}-} & {\footnotesize{}39.8 / 61.2 / 85.7}\tabularnewline
\hline 
{\footnotesize{}D2-Net (single)} & {\footnotesize{}-} & {\footnotesize{}44.9 / 66.3 / \textbf{88.8}}\tabularnewline
\hline 
{\footnotesize{}D2-Net (multi)} & {\footnotesize{}-} & {\footnotesize{}44.9 / 64.3 / \textbf{88.8}}\tabularnewline
\hline 
{\footnotesize{}R2D2 (web)} & {\footnotesize{}-} & {\footnotesize{}43.9 / 61.2 / 77.6}\tabularnewline
\hline 
{\footnotesize{}R2D2 (aachen day)} & {\footnotesize{}-} & {\footnotesize{}45.9 / 65.3 / 86.7}\tabularnewline
\hline 
{\footnotesize{}ContextDesc} & {\footnotesize{}-} & {\footnotesize{}\textbf{46.9} / 65.3 / 87.8}\tabularnewline
\hline 
{\footnotesize{}UR2KID (single)} & {\footnotesize{}-} & {\footnotesize{}\textbf{46.9} / \textbf{67.3} / \textbf{88.8}}\tabularnewline
\hline 
\multicolumn{3}{c}{{\footnotesize{}Protocol 2 (Global retrieval for candidate ranking)}}\tabularnewline
\hline 
{\footnotesize{}ESAC (50 experts)} & {\footnotesize{}42.6 / 59.6 / 75.5} & {\footnotesize{}3.1 / 9.2 / 11.2}\tabularnewline
\hline 
{\footnotesize{}AS} & {\footnotesize{}57.3 / 83.7 / \textbf{96.6}} & {\footnotesize{}19.4 / 30.6 / 43.9}\tabularnewline
\hline 
{\footnotesize{}NV+Superpoint} & {\footnotesize{}79.7 / 88.0 / 93.7} & {\footnotesize{}40.8 / 56.1 / 74.5}\tabularnewline
\hline 
{\footnotesize{}HF-Net} & {\footnotesize{}75.7 / 84.3 / 90.9} & {\footnotesize{}40.8 / 55.1 / 72.4}\tabularnewline
\hline 
{\footnotesize{}NV+D2-Net (single)} & {\footnotesize{}79.7 / \textbf{89.3} / 94.8} & {\footnotesize{}41.8 / 63.3 / 81.6}\tabularnewline
\hline 
{\footnotesize{}UR2KID (single)} & {\footnotesize{}\textbf{79.9} / 88.6 / 93.6 } & {\footnotesize{}\textbf{45.9} / \textbf{64.3} / \textbf{83.7}}\tabularnewline
\hline 
\end{tabular}{\small\par}
\par\end{centering}
\caption{\label{tab:aachen} Aachen day-night comparison with the localization threshold for day: (0.25m, 2$^{\circ}$ ) / (0.5m, 5$^{\circ}$ ) / (5m, 10$^{\circ}$ ), and night: (0.5m, 2$^{\circ}$ ) / (1m, 5$^{\circ}$ ) / (5m, 10$^{\circ}$ ). Our method is able to achieve top results in both scenarios.}
\end{table}

\begin{figure}
\begin{centering}
\includegraphics[scale=0.3]{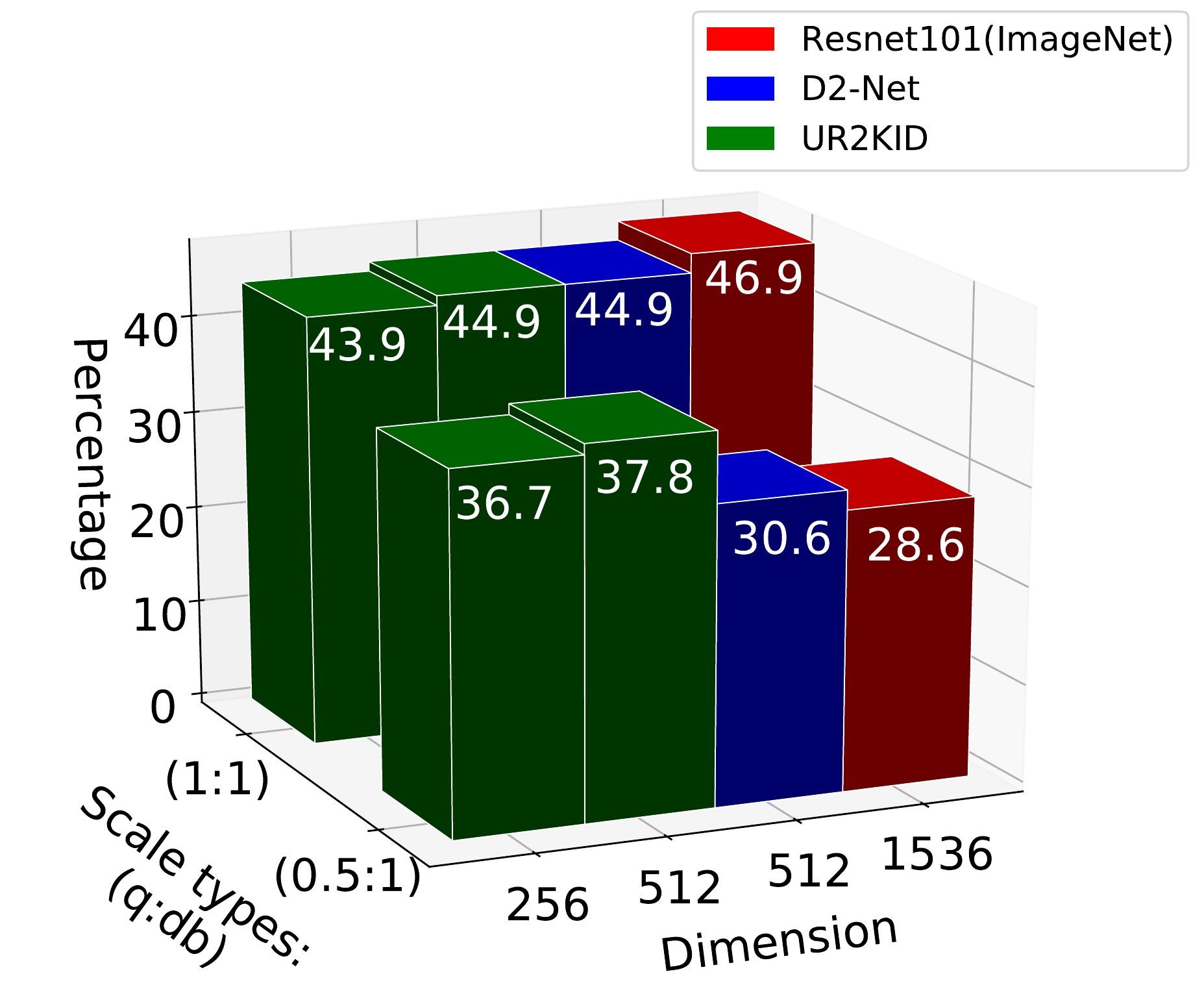} 
\par\end{centering}
\caption{\label{fig:scale} Scale comparison for Aachen dataset with localization threshold at night: (0.5m, 2$^{\circ}$ ). The query versus database scales are shown as $(q,db)$. We can observe that our method is very robust against scale changes.}
\end{figure}

\begin{table*}
\small
\begin{centering}
{\footnotesize{}{}}%
\begin{tabular}{|c|c|c|c|c|c|c|c|c|c|c|c|c|c|c|}
\hline 
\multicolumn{2}{|c|}{Dim} & \multicolumn{1}{c|}{} & \multicolumn{3}{c|}{$r_{q}:r_{db}=1:1$} & \multicolumn{3}{c|}{$r_{q}:r_{db}=0.5:1$} & \multicolumn{3}{c|}{$r_{q}:r_{db}=0.5:0.5$} & \multicolumn{3}{c|}{$r_{q},r_{db}\in\left[0.25,1\right]$}\tabularnewline
\hline 
\multicolumn{15}{|c|}{UR2KID (TS): teacher detect / student desc}\tabularnewline
\hline 
B2  & B3  & Fix  & 0.5m  & 1m  & 5m  & 0.5m  & 1m  & 5m  & 0.5m  & 1m  & 5m  & 0.5m  & 1m  & 5m\tabularnewline
\hline 
256  & 256  & {\footnotesize{}{}$\checkmark$}  & \textbf{46.9}  & 67.3  & \textbf{88.8}  & \textbf{31.6}  & 49.0  & 66.3  & 35.7  & 56.1  & 78.6  & 35.7  & 54.1  & 74.5\tabularnewline
\hline 
256  & 256  & -  & 44.9  & 66.3  & \textbf{88.8}  & 28.6  & 49.0  & 67.3  & 36.7  & 63.3  & 81.6  & 37.8  & \textbf{62.2}  & \textbf{83.7}\tabularnewline
\hline 
128  & 128  & {\footnotesize{}{}$\checkmark$}  & 44.9  & 68.4  & \textbf{88.8}  & 27.6  & 44.9  & 63.3  & 34.7  & 51.0  & 76.5  & 30.6  & 51.0  & 69.4\tabularnewline
\hline 
128  & 128  & -  & 41.8  & 65.3  & 86.7  & 28.6  & 43.9  & 61.2  & 34.7  & 53.1  & 77.6  & 34.7  & 51.0  & 71.4\tabularnewline
\hline 
\hline 
\multicolumn{15}{|c|}{UR2KID (SS): student detect / student desc}\tabularnewline
\hline 
256  & 256  & {\footnotesize{}{}$\checkmark$}  & 44.9  & \textbf{69.4 } & \textbf{88.8}  & \textbf{37.8}  & \textbf{52.0}  & \textbf{73.5}  & \textbf{41.8}  & \textbf{64.3}  & \textbf{86.7}  & \textbf{39.8}  & \textbf{62.2}  & 80.6\tabularnewline
\hline 
256  & 256  & -  & 43.9  & 67.3  & 87.8  & 34.7  & 46.9  & \textbf{73.5}  & 38.8  & 60.2  & 84.7  & 35.7  & \textbf{62.2}  & 79.6\tabularnewline
\hline 
128  & 128  & {\footnotesize{}{}$\checkmark$}  & 43.9  & 66.3  & 86.7  & 36.7  & 51.0  & \textbf{73.5}  & 37.8  & 59.2  & 84.7  & 38.8  & 59.2  & 82.7\tabularnewline
\hline 
128  & 128  & -  & 42.9  & 66.3  & 87.8  & 33.7  & 49.0  & 69.4  & 40.8  & 56.1  & 82.7  & 37.8  & 59.2  & 79.6\tabularnewline
\hline 
\hline 
\multicolumn{3}{|c|}{d2-net (512)} & 44.9  & 66.3  & \textbf{88.8}  & 30.6  & 45.9  & 65.3  & 36.7  & 58.2  & 80.6  & 38.8  & 55.1  & 80.6\tabularnewline
\hline 
\end{tabular}
\par\end{centering}
\begin{centering}
 
\par\end{centering}
\caption{\label{tab:ablation} Ablation study for mapped student feature output. 
The "Fix" option means the weights before ResNet block2 (B2) and block3 (B3) are frozen or not. 
The "Dim" means the final mapped output local descriptor dimension. }
\end{table*}

\begin{table}
\begin{centering}
\begin{tabular}{c|c|c}
\hline 
 & Oxf5k & Par6k\tabularnewline
\hline 
\hline 
SIFT & 51.64 & 52.23\tabularnewline
\hline 
Geodesc & 54.98 & 55.02\tabularnewline
\hline 
Contextdesc & 65.03 & 64.53\tabularnewline
\hline 
LIFT & 54.0 & 53.6\tabularnewline
\hline 
\hline 
NetVLAD & 71.6 & 79.7\tabularnewline
\hline 
\textbf{UR2KID (megadepth)} & 82.03 & 91.94\tabularnewline
\hline 
GeM & 88.17 & 92.6\tabularnewline
\hline 
DAME & 88.24 & 93.0\tabularnewline
\hline 
\textbf{UR2KID (sfm120k)} & 88.75 & 93.0\tabularnewline
\hline 
DELF & 90.0 & 95.7\tabularnewline
\hline 
\end{tabular}
\par\end{centering}
\caption{\label{tab:retrieval} Image retrieval mean average precision (mAP) comparison for Oxford5k and Paris6k datasets. Our method not only achieves state-of-the-art on the previous matching benchmarks, but also has competitive performance in global landmark retrieval.}

\end{table}

\textbf{Testing} To evaluate the performance of the local matching pipeline, and of the global image retrieval, we examine several different datasets for testing a set of representative scenarios.

\textbf{(a) Hpatches }dataset \cite{balntas2019hpatches} is the most well-known image matching dataset for identifying the matching robustness against different illumination and viewpoint changes. 
We compute the MMA (mean matching accuracy) as indicated in the D2-Net \cite{dusmanu2019d2} along with other state-of-the-art methods for verifying the performance of our method.
\textbf{ b) Aachen Day-Night} dataset contains 98 night-time query images in the testing dataset, along with 20 relevant images in day-time with known ground truth camera poses. 
After the keypoint and descriptor extraction based on 4479 database images, COLMAP pipeline is adopted for the 3D reconstruction.
We follow the evaluation protocol from \cite{sattler2018benchmarking} and D2-Net \cite{dusmanu2019d2}, and the percentage of the queries localized within a given error bound on the estimated camera position and orientation is reported.
Both keypoint detection and descriptor matching contribute to the camera localization task. \textbf{c) Oxford5k, Paris6k} dataset which consist of $55$ query images with bounding boxes, with the images exhibiting significant background noise such as people, trees etc. 
5,062 building images are captured in Oxford, and 6,412 images of landmarks in Paris. Following the GeM setting, the global descriptor is tested in both datasets.

\subsection{Hpatches dataset}
Similar to the previous state-of-the-art methods in patch descriptor or matching pipeline, we evaluate the proposed UR2KID on the Hpatches dataset. 
In Figure \ref{fig:hpatches}(a), we demonstrate the comparison over multiple state-of-the-art methods in both illumination changes and viewpoint changes. 
Considering that our training does not depend on any special augmentation and pointwise correspondence ground-truth supervision, it is remarkable to see UR2KID outperform other methods such as Superpoint \cite{detone2018superpoint} and Lf-Net \cite{ono2018lf} by a margin in illumination changes. 
Notice that compared to D2-Net \cite{dusmanu2019d2}, which is the most relevant method comparing to ours, a clear advantage can be observed in the experimental results considering our method is only based on single scale while D2-Net \cite{dusmanu2019d2} uses multiple scale inputs for boosting the performance in viewpoint changes.
Among the state-of-the-art methods, DELF \cite{Noh2017Large} is the most extreme case considering it's the most robust one against illumination change while it is also the most vulnerable one against viewpoint changes. Note that R2D2 \cite{revaud2019r2d2} achieves very high performance on the viewpoint changes with very large amount of scales ($8\sim9$ scales) which consume linear amount of time against the sampled scales. On the other hand, our method can outperform them on the single scale constraint.

\textbf{Ablation studies}  In Figure \ref{fig:hpatches}(b), we show the ablation study among different group choices as described in Section \ref{sec:gc-dad}. Generally speaking, increasing the keypoint number by tuning the threshold is capable of generating more correspondence candidates. 
However, as shown in Figure \ref{fig:Grouping-channel-keypoint}, there are a lot of portion of the keypoints detected by D2-Net which are not on the target buildings and cause false matches. For a fair comparison, we compare DAD and GC-DAD with different parameters on the same trained network, ResNet101, based on UR2KID pipeline and the keypoint numbers are also shown in Figure \ref{fig:hpatches}(b).
We can observe that Our method with increased number of groups  is able to capture the different concepts across the channel while maintaining good correspondence rate.

\subsection{Aachen Day-Night dataset}

Comparing to a pure local matching benchmark such as HPatches, a localization benchmark such as Aachen-day-night \cite{sattler2018benchmarking} involves more steps including local matching, geometric verification, triangulation, bundle adjustment, and solving the PnP problem. 
A structure-from-motion model is built based on the database images while the goal of the query is to recover the camera pose from the given query image. 
It's a more challenge standard for the sparse local matching based methods because local matching is the first step of the SfM pipeline and the error will be amplified after going through the aforementioned steps. 

In Table \ref{tab:aachen}, we demonstrate two different evaluation protocols. 
The first one is supported by the ground-truth query-database pair candidates. 
In this case, it is only required to match the local descriptor sets between the query image and the provided candidates from the database images. 
The proposed UR2KID is the only method that optimizing the local matches without pointwise supervision and the best performance is obtained, with a single scale.    

The second protocol is based on global descriptor retrieval and there is no given query-database pair. 
We follow the similar setting as described in \cite{dusmanu2019d2, sattler2018benchmarking, sarlin2019coarse}, and the top-20 retrieved images are consider as the query-database pairs for the local matching. 
Using NetVLAD with D2-Net single scale for the localization, the performance degrades a little comparing to protocol 1 due to the fact that retrieved images may not form the optimal query-database pair comparing to the ground-truth. 
ESAC \cite{brachmann2019expert} is the family of unifying partial pipeline of the localization process. Nevertheless, it still suffers from the training data overfitting similar to PoseNet \cite{Kendall2015PoseNet} and DSAC \cite{Brachmann2017DSAC}. 
By using the global descriptor from our multi-task framework, UR2KID is able to provide both retrieval candidates and the local matching for the localization task. 
Our method is comparable in day case and outperforms the other methods at night.

\textbf{Ablation studies} In order to identify the strength of the proposed method, we examine ablation studies in Figure \ref{fig:scale} and Table \ref{tab:ablation} based on the protocol 1 in Aachen dataset. 

Despite the localization task is more challenging comparing to the pure matching benchmark, the dataset design is still non-realistic enough considering we cannot be sure about the target building size is large enough in the taken query image for localization. 
Therefore, we make different ratios between the query and the databased images into $(r_{q}:r_{db})=(1:1)$ and $(r_{q}:r_{db})=(0.5:1)$ in Figure \ref{fig:scale}. As we previously stated, the ResNet101 is indeed very robust in  the localization task with our GC-DAD paradigm with high dimensional local descriptors.
However, the performance degrades severely when it comes to $(r_{q}:r_{db})=(0.5:1)$ with massive scale changes and similar degradation is observed in D2-Net. 
On the other hand, our method is not only robust in $(r_{q}:r_{db})=(1:1)$, and the performance in $(r_{q}:r_{db})=(0.5:1)$ case is also much higher comparing the others even with much lower dimension descriptor (e.g. 256, 512).

In Table \ref{tab:ablation}, more details about the training process along with different scale changes variants are compared between our UR2KID and D2-Net considering they are highly related. 
Four different ratio combinations are discussed. $(r_{q}:r_{db})=(1:1),(0.5:1),(0.5:0.5)$ and a random ratio combination between $[0.25,1]$. 
We examine different mapping dimension (256,128 per block), detector choices (teacher or student), and the frozen weight choice during training (freeze block2 and block3 or not). 
Among those combinations, we can see that the trained student detector with student descriptor achieved the best results in the different variants, while the teacher detector achieves the best results when there is no scale change between query and database images. 
In every case, freezing the weight during training is the best option which indicates that fine-tuning the mapping layer only is enough for learning the low dimensional descriptor. 
On the other hand, D2-Net performs poorly comparing to ours, giving $5\%-7\%$ worse results for severe scale changes.

\subsection{Oxford5k, Paris6k dataset}
In Table \ref{tab:retrieval}, we compare different methods with our global descriptor in the retrieval framework. 
Based on the previous ablation study we know that the localization performance can be optimized when we freeze the weights and fine-tune the mapping layer only, and the retrieval task is also trained based on such setting. 
Our method is better than most visual word retrieval method such as ContextDesc \cite{luo2019contextdesc}, GeoDesc \cite{luo2018geodesc}, and LIFT \cite{yi2016lift} when trained on the megadepth dataset. However, the performance is only comparable or lower than the retrieval task specific methods such as GeM \cite{Radenovi2018Fine}, DAME \cite{yang2019dame} and DELF \cite{Noh2017Large}. As shown in Table~\ref{tab:retrieval}, the performance is greatly improved by using SfM120k dataset as suggested by GeM \cite{Radenovi2018Fine}. In Oxford5k dataset, there are significant scale differences of the landmarks between the query image and the dataset reference image as discussed in \cite{yang2019dame} which is consistent with the SfM120k dataset, while Paris6k suffers less from that point of view. Megadepth is suitable for matching as in D2-Net \cite{dusmanu2019d2} by estimating the overlap ratio, but it is not suitable for global retrieval training.

These competitive results indicate that it is possible to embed local and global tasks into one single network.


%% file: conclusion.tex
\section{Conclusion}

We propose a multi-task framework for unifying global context along with local descriptors suitable for both retrieval and local matching tasks. The method exploits full image pairs instead of pointwise supervision during training while exhibiting state-of-the-art matching and localization results. 
In addition, we explore how significant scale changes affect the localization benchmark and identify that previous state-of-the-art  methods are vulnerable against scale changes between query and database. Compared to other methods, our method is more robust against illumination and viewpoint changes, day-night shifting, and scale changes. 
As many new strong backbones (e.g. EfficientNet \cite{tan2019efficientnet}) have been recently developed, we believe that our work is important for simplifying multi-task methods into a single network with very weak label information and can inspire similar future studies.

\paragraph*{Acknowledgement}
This work was supported partially by the Computer Vision Lab of Tohoku University.